\documentclass[runningheads, envcountsame, a4paper]{llncs}

\usepackage{xcolor}
\usepackage{balance}
\usepackage{subfig}
\usepackage[pdftex]{graphicx}
\usepackage{epstopdf}
\usepackage{graphics}
\usepackage{multirow}
\usepackage[hyphens]{url}
\usepackage{times}
\usepackage{booktabs} 
\usepackage[misc]{ifsym}
\usepackage{microtype}

\begin{document}

\toctitle{Learning Cheap and Novel Flight Itineraries}
\tocauthor{Dmytro~Karamshuk and David~Matthews}
\title{Learning Cheap and Novel Flight Itineraries}
\author{Dmytro Karamshuk [\Letter] \and David Matthews}

\institute{Skyscanner Ltd, United Kingdom \\
\email{Dima.Karamshuk@skyscanner.net} \email{David.Matthews@skyscanner.net} }
\maketitle

\begin{abstract}

We consider the problem of efficiently constructing cheap and novel round trip flight itineraries by combining legs from different airlines. We analyse the factors that contribute towards the price of such itineraries and find that many result from the combination of just 30\% of airlines and that the closer the departure of such itineraries is to the user's search date the more likely they are to be cheaper than the tickets from one airline. We use these insights to formulate the problem as a trade-off between the recall of cheap itinerary constructions and the costs associated with building them.

We propose a supervised learning solution with location embeddings which achieves an AUC=80.48, a substantial improvement over simpler baselines. We discuss various practical considerations for dealing with the staleness and the stability of the model and present the design of the machine learning pipeline. Finally, we present an analysis of the model's performance in production and its impact on Skyscanner's users.

\end{abstract}

\section{Introduction}

\begin{figure*}
\centering
\includegraphics[width = 0.8\columnwidth]{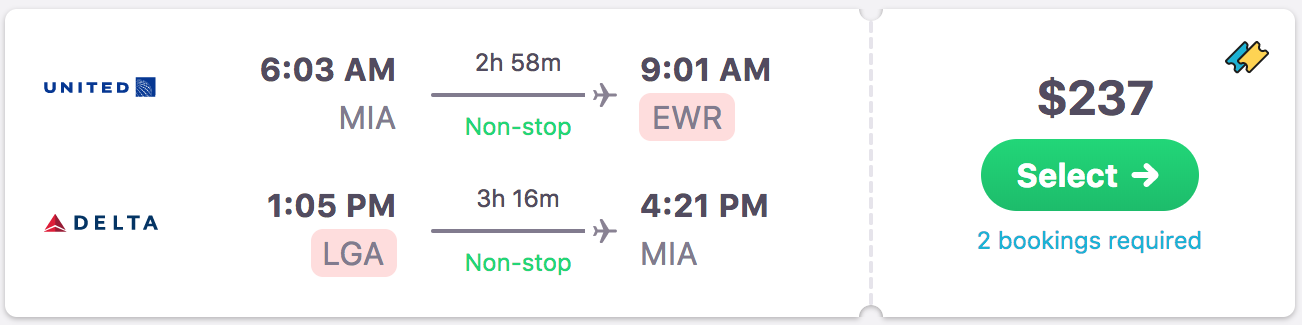}
\caption{Example of a combination flight itinerary in Skyscanner's search results.}
\label{fig:mashup_itinerary}
\end{figure*}

Different strategies are used by airlines to price round trip tickets. Budget airlines price a complete round trip flight as the sum of the prices of the individual outbound and inbound journeys (often called flight legs). This contrasts with traditional, national carrier, airlines as their prices for round trip flights are rarely the sum of the two legs. Metasearch engines, such as Skyscanner\footnote{https://www.skyscanner.net/}, can mix outbound and inbound tickets from different airlines to create combination itineraries, e.g., flying from Miami to New York with United Airlines and returning with Delta Airlines (Fig.~\ref{fig:mashup_itinerary})\footnote{Our constructions contrast with those built through interlining which involve two airlines combining flights on the same leg of a journey organised through a commercial agreement.}. Such combinations are, for a half of search requests, cheaper than the round trip tickets from one airline. 

A na\"ive approach to create such combinations with traditional airlines, requires an extra two requests for prices per airline, for both the outbound and the inbound legs, on top of the prices for complete round trips. These additional requests for quotes is an extra cost for a metasearch engine. The cost, however, can be considerably optimized by constructing only the combinations which are competitive against the round trip fares from airlines. 

To this end, we aim to predict price competitive combinations of tickets from traditional airlines given a limited budget of extra quote requests. Our approach is as follows.

Firstly, we analyse a data set of 2.3M search queries from 768K Skyscanner's users, looking for the signals which impact the competitiveness of combination itineraries in the search results. We find the that the vast majority of competitive combination itineraries are composed of only 30\% of airlines and are more likely to appear in the searches for flights departing within days of the user's search.

Secondly, we formulate the problem of predictive itinerary construction as a trade-off between the computation cost and resulting coverage, where the cost is associated with the volume of quote requests the system has to make to construct combination itineraries, and the coverage represents the model's performance in finding all such itineraries that are deemed price competitive. To the best of our knowledge this is the first published attempt to formulate and solve the problem of constructing flight itineraries using machine learning.

Thirdly, we evaluate different supervised learning approaches to solve this problem and propose a solution based on neural location embeddings which outperforms simpler baselines and achieves an AUC=80.48. We also provide an intuition on the semantics of information that such embedding methods are able to learn.

Finally, we implement and deploy the proposed model in a production environment. We provide simple guidance for achieving the right balance between the staleness and stability of the production model and present the summary of its performance.

\section{Data set}
\label{sec:dataset}

To collect a dataset for our analysis we enabled the retrieval of both outbound and inbound prices for all airlines on a sample of 2.3M Skyscanner search results for round trip flights in January 2018. We constructed all possible combination itineraries and recorded their position in the ranking of the cheapest search results, labelling them competitive, if they appeared in the cheapest ten search results, or non-competitive otherwise\footnote{Skyscanner allows to rank search results by a variety of other parameters apart from the cheapest. The analysis of these different ranking strategies is beyond the scope of this paper.}. This resulted in a sample of 16.9M combination itineraries (both competitive and non-competitive) for our analysis, consisting of 768K users searching for flights on 147K different routes, i.e., origin and destination pairs.

Our analysis determined that the following factors contribute towards a combination itinerary being competitive.

\subsection{Diversity of airlines and routes} 
\begin{figure*}
\centering
\includegraphics[width=0.65\columnwidth]{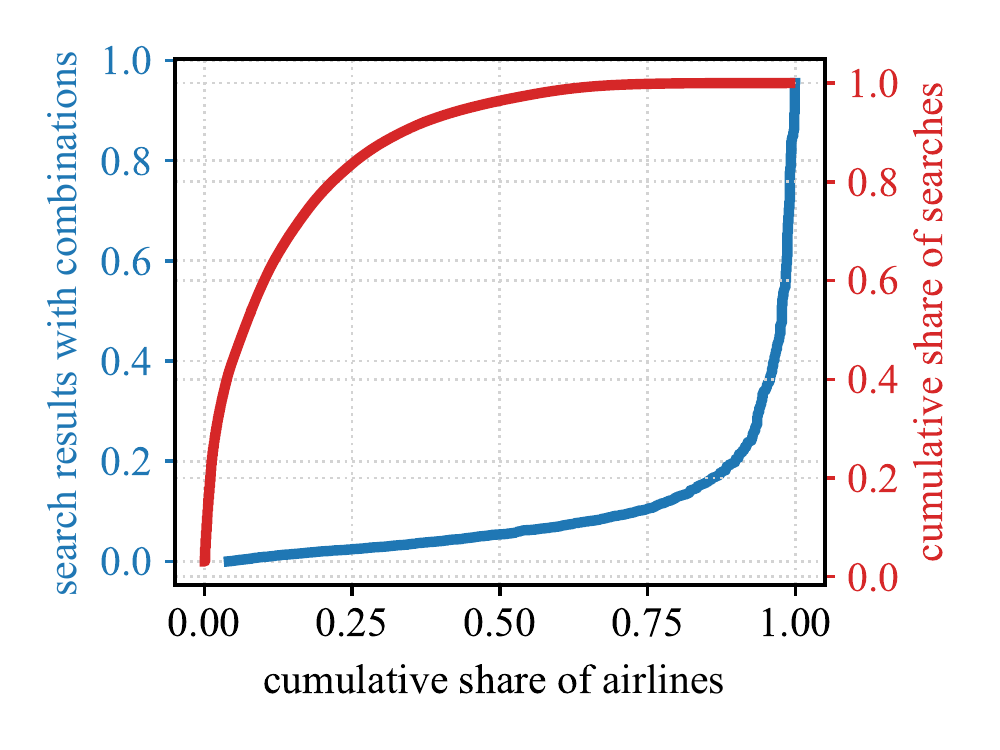}
\caption{Search results with competitive combinations across different airlines. The cumulative share of all search results (red) and search results with competitive combinations (blue) for top x\% of airlines (x-axis).}
\label{fig:mashup_partners_diversity}
\end{figure*}

We notice that the vast majority (70\%) of airlines rarely appear in a competitive combination itinerary (Fig.~\ref{fig:mashup_partners_diversity}), i.e., they have a less than 10\% chance of appearing in the top ten of search results. The popularity of airlines is highly skewed too. The top 25\% of airlines appear in 80\% of the search results whereas the remaining 75\% of airlines account for the remaining 20\%. We found no correlation between airlines' popularity and its ability to appear in a competitive combination itinerary.

\begin{figure*}
\centering
\includegraphics[width=0.65\columnwidth]{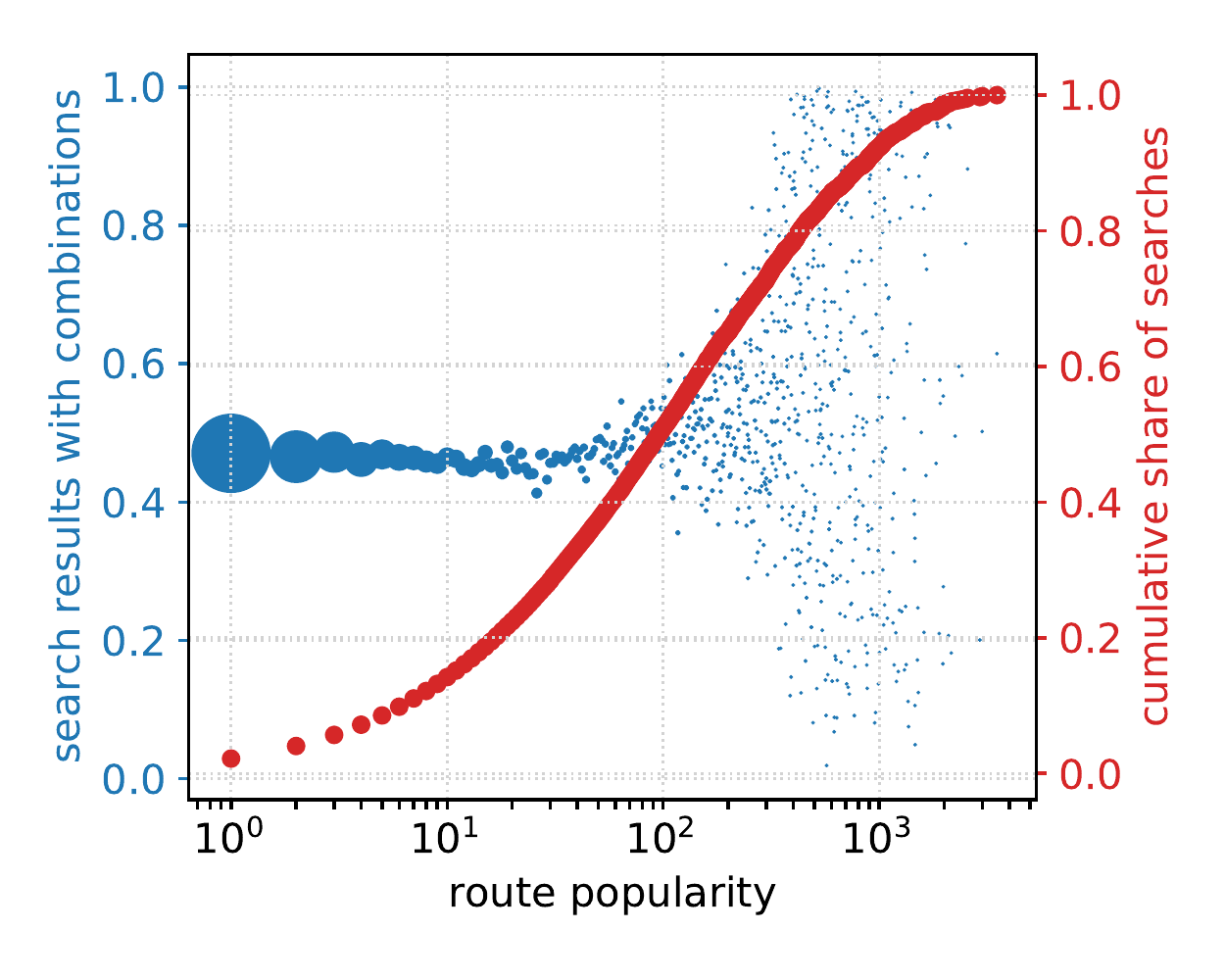}
\caption{Search results with competitive combinations across routes with different popularity. Red: the cumulative distribution function of the volume of searches across different origin and destination pairs (routes). Blue: the share of search results with competitive combinations (y-axis) on the routes of a given popularity (x-axis).}
\label{fig:mashup_prices_popularity}
\end{figure*}

The absence of a correlation with popularity is even more vividly seen in the analysis of combination performance on different search routes (Fig.~\ref{fig:mashup_prices_popularity}). The share of competitive combinations on unpopular and medium popular routes is rather stable ($\approx45\%$) and big variations appear only in the tail of popular routes. In fact, some of those very popular routes have almost a 100\% chance to have combination itineraries in the top ten results, whereas some other ones of a comparable popularity almost never feature a competitive combination itinerary.

This finding is in line with our modelling results in section \ref{sec:model} where we observe that the popularity of a route or an airline is not an indicative feature to predict price competitiveness of combination itineraries. We therefore focus on a small number of airlines and routes which are likely to create competitive combination itineraries. We explore different supervised learning approaches to achieve this in section \ref{sec:model}.

\subsection{Temporal patterns} 
\begin{figure*}
\centering
\includegraphics[width=0.65\columnwidth]{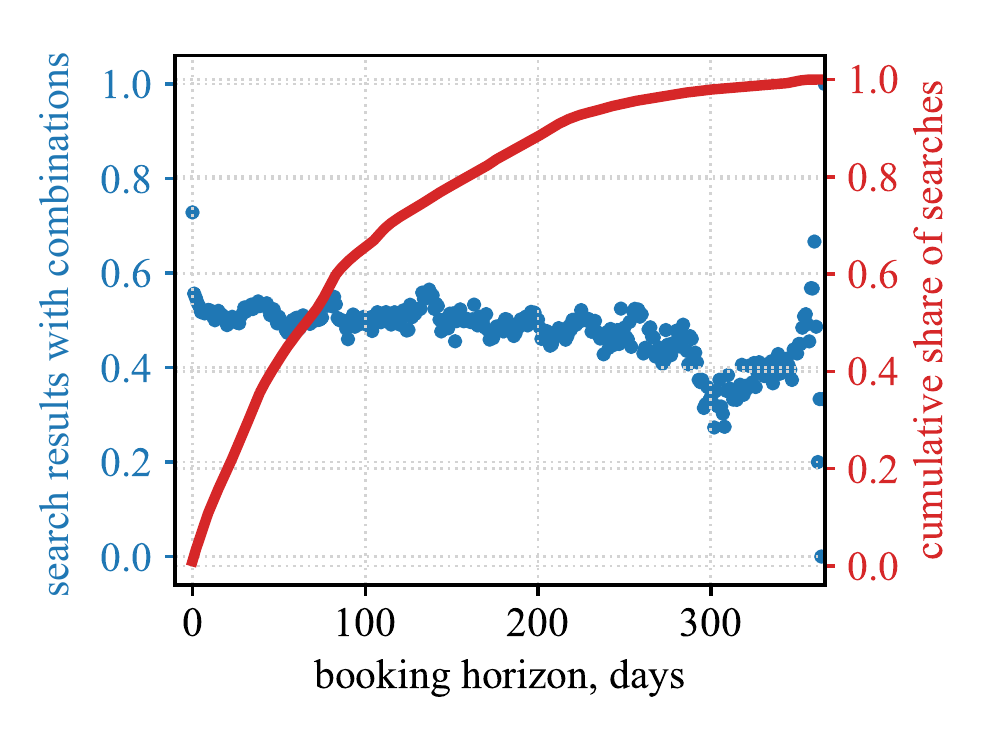}
\caption{Search results with competitive combinations across different days between search and departures (booking horizon). Red: the cumulative distribution function of the booking horizon. Blue: the share of search results with competitive combinations (y-axis) for a given booking horizon (x-axis).}
\label{fig:mashup_horizon}
\end{figure*}

We also analyse how the days between search and departure (number of days before departure) affects the competitiveness of combinations in the top ten of search results (Fig.~\ref{fig:mashup_horizon}). We find that combination itineraries are more likely to be useful for searches with short horizons and gradually become less so as the days between search and departure increases. One possible explanation lies in the fact that traditional single flight tickets become more expensive as the departure day approaches, often unequally so across different airlines and directions. Thus, a search for a combination of airlines on different flight legs might give a much more competitive result. This observation also highlights the importance to consider the volatility of prices as the days between search and departure approaches, the fact which we explore in building a production pipeline in section \ref{sec:production}.

 \section{Predictive construction of combination itineraries}
\label{sec:model}

Only 10\% of all possible combination itineraries are cheap enough to appear in the top ten cheapest results and therefore be likely to be seen by the user. The difficulty is in the fact that the cost of enabling combinations in Skyscanner search results is proportional to the volume of quote requests required to check their competitiveness. In this section we formulate the problem of predictive combination itinerary construction where we aim to train an algorithm to speculatively construct only those combinations which are likely to be competitive and thus to reduce the overall cost associated with enabling combinations in production.

\subsection{Problem formulation}

We tackle the predictive combination itinerary construction as a supervised learning problem where we train a classifier $F(Q, A, F) \rightarrow \{True, False\}$ to predict whether any constructed combination itinerary in which airline $A$ appears on the flight leg $F$, either outbound or inbound, will yield a competitive combination itinerary in the search results for the query $Q$. The current formulation is adopted to fit in Skyscanner's current pricing architecture which requires an advance decision about whether to request a quote from airline $A$ on a leg $F$ for a query $Q$. To measure the predictive performance of any such classifier $F(Q, A, F)$ we define the following metrics:

Recall or coverage is measured as a share of competitive itineraries constructed by the classifier $F(X)$, more formally:

\begin{equation}
Recall@10 = \frac{|L^{@10}_{pred} \cap L^{@10}_{all}|}{|L^{@10}_{all}|}
\end{equation}

where $L^{@10}_{pred}$ is the set of competitive combination itineraries constructed by an algorithm and $L^{@10}_{all}$ is the set of all possible competitive combination itineraries. In order to estimate the latter we need a mechanism to sample the ground truth space which we discuss in section \ref{sec:production}.

Quote requests or cost is measured in terms of all quote requests required by the algorithm to construct combination itineraries, i.e.:

\begin{equation}
Quote \   Requests = \frac{|L_{pred}|}{|L_{all}|}
\end{equation}

where $L_{all}$ - is the set of all possible combination itineraries constructed via the ground truth sampling process. Note that our definition of the cost is sometimes also named as predictive positive condition rate in the literature.

The problem of finding the optimal classifier $F(Q, A, F)$ is then one of finding the optimal balance between the recall and quote requests. Since every algorithm can yield a spectrum of all possible trade-offs between the recall and the quote requests we also use the area under the curve (AUC) as an aggregate performance metric.

\subsection{Models}

We tried several popular supervised learning models including logistic regression, multi-armed bandit and random forest. The first two algorithms represent rather simple models which model a linear combination of features (logistic regression) or their joint probabilities (multi-armed bandit). In contrast, random forest can model non-linear relations between individual features and exploits an idea of assembling different simple models trained on a random selection of individual features. We use the scikit-learn\footnote{http://scikit-learn.org/} implementation of these algorithms and benchmark them against:

\paragraph{Popularity baseline} We compare the performance of the proposed models against a na\"{i}ve popularity baseline computed by ranking the combinations of (origin, destination, airline) by their popularity in the training set and cutting-off the top K routes which are estimated to cumulatively account for a defined share of quote requests. We note that this is also the model which was initially implemented in the production system.

\begin{figure}
\begin{minipage}{.34\columnwidth}
\resizebox{\columnwidth}{!}{%
\begin{tabular}{ |c|c| }
  \hline
  \textbf{Model} & \textbf{AUC}\\ \hline
  popularity & 51.76\\
  logistic regression & 75.69\\
  multi-armed bandit & 77.68\\
  random forest & 80.37 \\
  oracle & 96.60 \\
  \hline
\end{tabular}}
\end{minipage}
\begin{minipage}{.65\columnwidth}
\resizebox{\columnwidth}{!}{%
\includegraphics{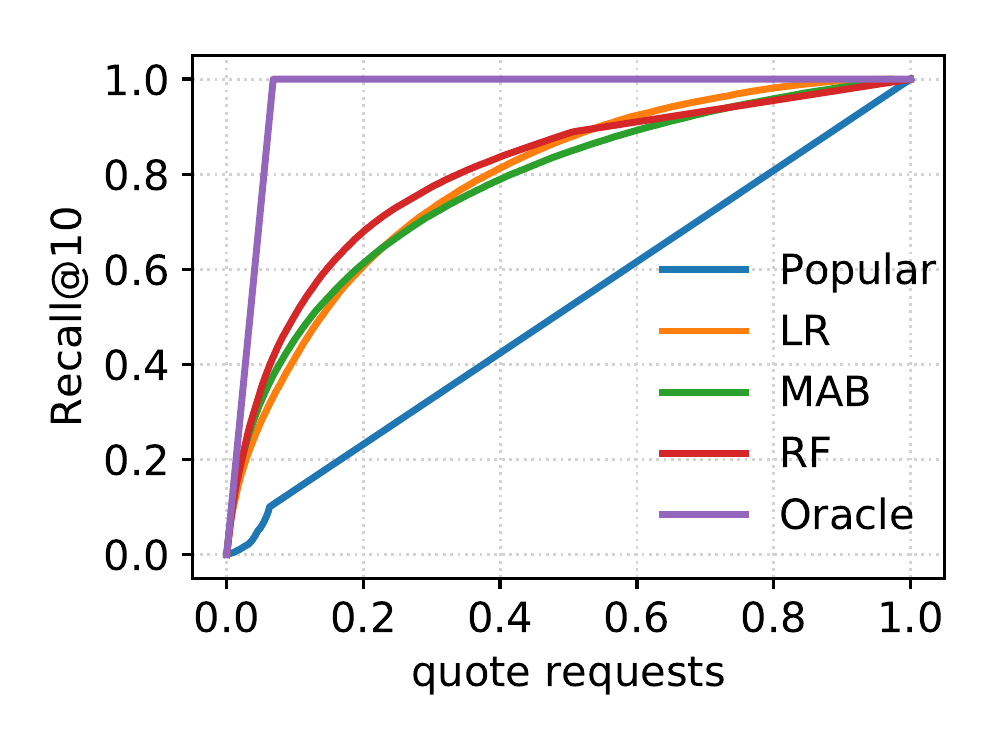}}
\end{minipage}
\caption{Performance of different supervised learning models (logistic regression (LR), nearest neighbour (NN), multi-armed bandit (MAB) and random forest (RF)) bench-marked over a na\"{i}ve popularity baseline (popular) and the upper-bound performance attainable with a perfect knowledge of the future (oracle).}
\label{fig:model_luc_recall}
\end{figure}

\paragraph{Oracle upper-bound} We also define an upper-bound for the prediction performance of any algorithm by considering an oracle predictor constructed with the perfect knowledge of the future, i.e., the validation data set. The aim of the oracle predictor is to estimate the upper-bound recall of competitive combinations achieved with a given budget of quote requests.

\paragraph{Results} From Fig.~\ref{fig:model_luc_recall} we observe that all proposed supervised models achieve a superior performance in comparison to the na\"{i}ve popularity baseline (AUC = 51.76\%), confirming our expectations from section \ref{sec:dataset} that popularity alone cannot explain competitiveness of combinations itineraries. Next, we notice that the random forest model outperforms other models and achieves an AUC = 80.37\%, a large improvement from the second best performing model (AUC = 77.68\%). At the same time, the results of our best performing model still lag behind the oracle predictor which achieves 100\% recall with as little as 10\% of total cost or AUC = 96.60\%. In order to improve the performance of our best model even further in the following section we focused on experimenting with the representation of the feature space and more specifically the representation of location information identified as the most important predictor across all experiments.

\subsection{Location representations}
\label{sec:embeddings}

This section describes different approaches we tried to more richly represent location information.

\paragraph{Trace-based embeddings} In this approach we collected the histories of per-user searches in the training data set and built sequences of origin and destination pairs appearing in them. For instance, if a user searched for a flight from London to Barcelona, followed by a search from London to Frankfurt, followed by another one from Frankfurt to Budapest, then we will construct a sequence of locations [London, Barcelona, London, Frankfurt, Frankfurt, Budapest] to represent the user's history. We also filter out the users who searched for less than 10 flights in our data set and remove the duplicates in consecutive searches. We feed the resulting sequences into a Word2Vec algorithm~\cite{mikolov2013distributed}, treating each location as a word and each user sequence as a sentence. We end up with a representation of each origin and destination locations as vectors from the constructed space of location embeddings.

This approach is inspired by the results in mining distributed representations of categorical data, initially proposed for natural language processing~\cite{mikolov2013distributed}, but recently applied also for mining graph~\cite{perozzi2014deepwalk} and location data~\cite{pang2016deepcity}\cite{zhao2017geo}\cite{liu2016exploring}. Specifically, we tried the approach proposed in~\cite{pang2016deepcity} and~\cite{zhao2017geo}, but since the results were quite similar we only describe one of them.

\paragraph{Co-trained embeddings} In this alternate approach we train a neural network with embedding layers for origin and destination features, as proposed in~\cite{guo2016entity} and implemented in Keras embedding layers\footnote{https://keras.io/layers/embeddings/}. We use a six-layer architecture for our neural network where embedding layers are followed by four fully connected layers of 1024, 512, 256, 128 neurons with relu activation functions.

Note that the goal of this exercise is to understand whether we can learn useful representation of the location data rather than to comprehensively explore the application of deep neural networks as an alternative to our random forest algorithm which, as we discuss in section \ref{sec:production}, is currently implemented in our production pipeline. Hence, we focus on the representations we learn from the first layer of the proposed network.

\begin{table}
\centering
\begin{minipage}{.35\columnwidth}
\resizebox{\columnwidth}{!}{%
\begin{tabular}{ |c|c| }
  \hline
  \multicolumn{2}{|c|}{\textbf{London Heathrow}} \\ \hline
  \textbf{Airport} & \textbf{Similarity}\\ \hline
  Frankfurt am Main & 0.71\\
  Manchester & 0.69 \\
  Amsterdam Schipol & 0.62 \\
  Paris Charles de Gaulle & 0.62 \\
  London Gatwick & 0.61 \\
  \hline
\end{tabular}%
}
\end{minipage}
\begin{minipage}{.35\columnwidth}
\resizebox{0.85\columnwidth}{!}{%
\begin{tabular}{ |c|c| }
  \hline
  \multicolumn{2}{|c|}{\textbf{Beijing Capital }} \\ \hline
  \textbf{Airport} & \textbf{Similarity}\\ \hline
  Chubu Centrair & 0.91\\
  Taipei Taoyuan & 0.90 \\
  Seoul Incheon & 0.90 \\
  Miyazaki & 0.88 \\
  Shanghai Pudong & 0.88 \\
  \hline
\end{tabular}%
}
\end{minipage}
\caption{Examples of location embeddings for airports most similar to London Heathrow (left) and Beijing Capital (right) in the embedded feature space.}
\label{tab:embeddings}
\end{table}

\paragraph{Learned embeddings}

In Table~\ref{tab:embeddings} we present few examples of the location embeddings we learn with these proposed approaches. Particularly, we take few example airports (London Heathrow and Beijing Capital) and find other airports which are located in vicinity in the constructed vector spaces. The results reveal two interesting insights. Firstly, the resulting location embeddings look like they are capturing the proximity between the airports. The airports most closely located to London Heathrow and Beijing Capital are located in the western Europe and south-east Asia, correspondingly. Secondly, we notice that the algorithm is able to capture that London Heathrow is semantically much closer to transatlantic hubs such as Paris Charles de Gaulle, Amsterdam Schipol and London Gatwick rather than a geographically closer London Luton or London Stansted airports which are mainly focused on low-cost flights within Europe.

\begin{figure}
\begin{minipage}{.34\columnwidth}
\resizebox{\columnwidth}{!}{%
\begin{tabular}{ |c|c| }
  \hline
  \textbf{Model} & \textbf{AUC}\\ \hline
  One hot & 80.37\%\\
  Trace embeddings & 77.80\%\\
  DN embeddings & 80.48\%  \\
  Deep network (DN) & 82.67\%  \\
  Oracle & 96.60\% \\
  \hline
\end{tabular}}
\end{minipage}
\begin{minipage}{.65\columnwidth}
\resizebox{\columnwidth}{!}{%
\includegraphics{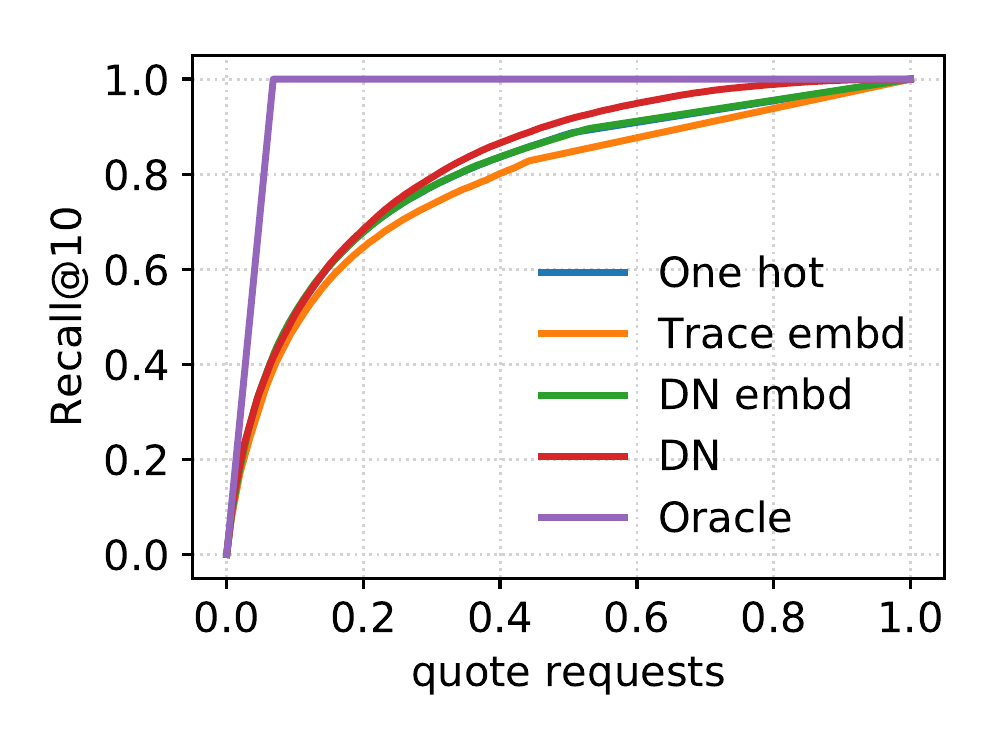}
}
\end{minipage}
\caption{Performance of the random forest model with different representations of origin and destination data (one hot encoding, trace-based embeddings, co-trained (DN) embeddings) and a neural network with embedding layers (DN).}
\label{fig:performance_encodings}
 \end{figure}

\subsection{Prediction performance} In Fig.~\ref{fig:performance_encodings} we compare the results of applying different location representations to the random forest algorithm proposed in the previous section. We use the random forest trained with one-hot representation as a baseline and compare it with: a) the random forest model trained with trace-based embeddings (orange curve) and b) the random forest trained with co-trained embeddings from the deep neural network model discussed early (green curve). In this latter approach we decouple the embedding layer from the rest of the layers in the neural network and use that as an input to our random forest model. We are able to assess how the embedding learned in the neural network can effectively represent the location data. Finally, we provide the results of the deep neural network itself for comparison (red curve).

The results of the model trained from trace-based embeddings performed worse than a baseline one-hot encoding, Fig.~\ref{fig:performance_encodings}. The random forest model with co-trained embeddings outperforms both results and achieves AUC = 80.48\%. The characteristic curves of the random forest model with one-hot encoding (blue curve) and co-trained embeddings (green curve) overlap largely in Fig.~\ref{fig:performance_encodings}, but a closer examination reveals a noticeable improvement of the latter in the area between 0 and 20\% and above 50\% of the quote request budget. One possible explanation behind these results might be that the embeddings we have trained from user-traces, in contrast to the co-trained embeddings, have been learning the general patterns in user-searches rather than optimising for our specific problem. 

We also notice that the performance of the deep neural network surpasses that of the random forest but any such comparison should also consider the complexity of each of the models, e.g., the number and the depth of the decision trees in the random forest model versus the number and the width of the layers in the neural network.

\section{Putting the model in production}
\label{sec:production}

\begin{figure}
\centering
\includegraphics[width = 0.65\columnwidth]{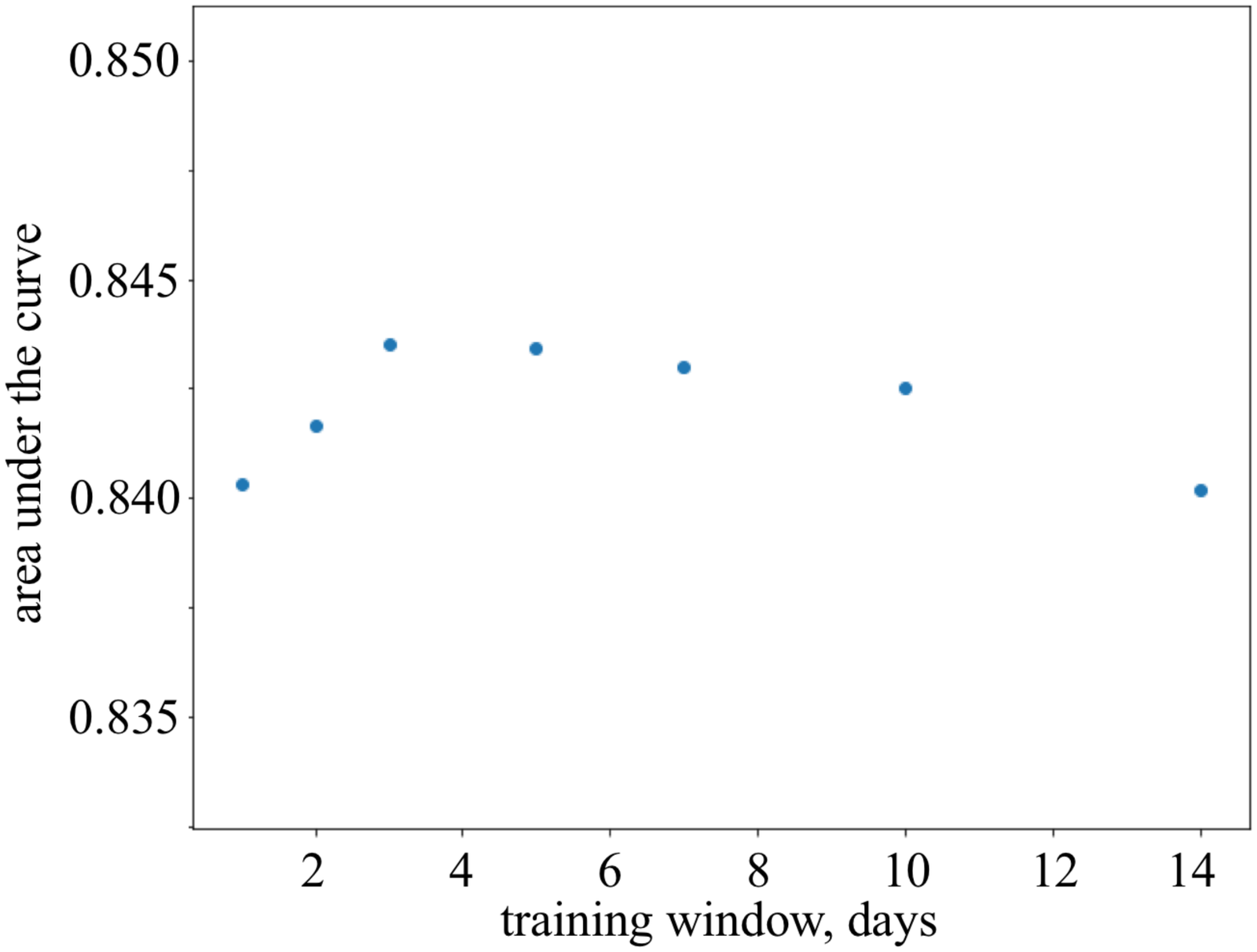}
\caption{The impact of the selected training window on the prediction performance of the model.}
\label{fig:training_window}
\end{figure}

\subsection{Model parameters}
\label{sec:temporal}

\paragraph{Training data window} To decide on how far back in time we need to look for data to train a good model we conduct an experiment where samples of an equivalent size are taken from each of the previous N days, for increasing values of N (Fig.~\ref{fig:training_window}). We observe that the performance of the model is initially increasing as we add more days into the training window, but slows down for N between $[3..7]$ days and the performance even drops as we keep increasing the size of the window further. We attribute this observation to the highly volatile nature of the flight fares and use a training window of 7 days to train the model in production.

\begin{figure}
\centering
\includegraphics[width = 0.55\columnwidth]{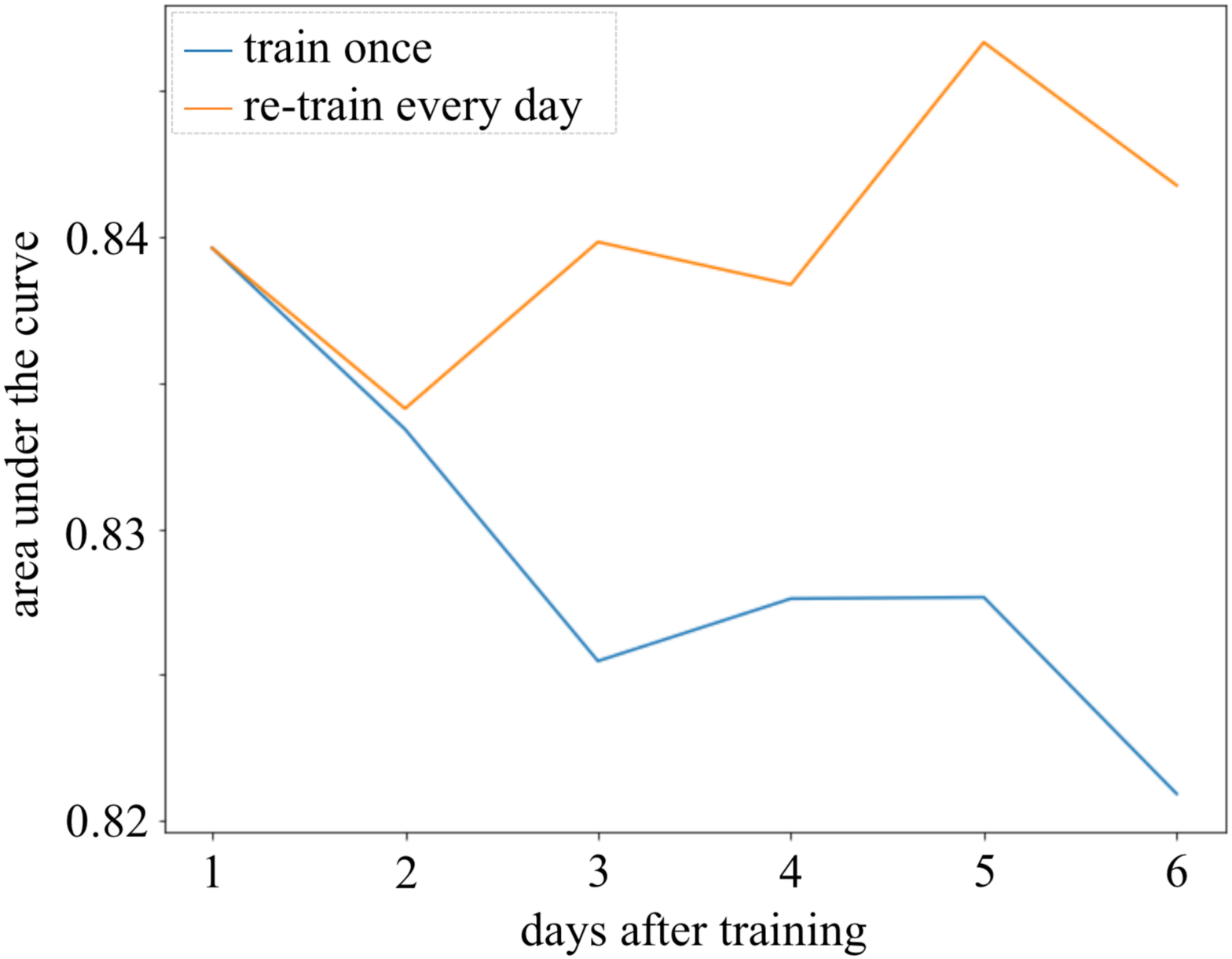}
\caption{Model staleness of the one-off trained model vs. the model retrained every day.}
\label{fig:temporal_validity}
\end{figure}

\paragraph{Model staleness} To decide how frequently to retrain the model in production we measure its staleness in an experiment (Fig.~\ref{fig:temporal_validity}). We consider a six day long period with two variants: when the model is trained once before the start of the experiment and when the model is retrained every single day. The results suggest, that the one-off trained model quickly stales by an average of ~0.3\% in AUC with every day of the experiment. The model retrained every single day, although also affected by daily fluctuations, outperforms the one-off trained model. This result motivates our decision to retrain the model every day.

\begin{figure}
\centering
\includegraphics[width = 0.65\columnwidth]{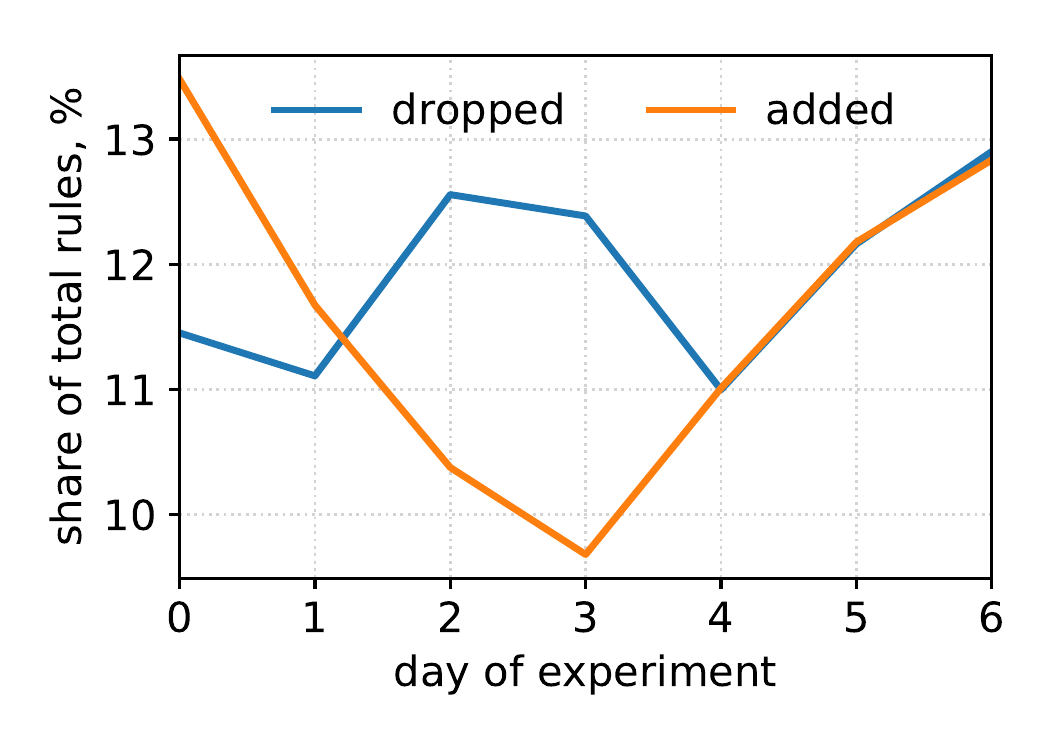}
\caption{Model stability, daily changes of (origin, destination, airline) rules inferred from the random forest model.}
\label{fig:temporal_stability}
\end{figure}

\paragraph{Model stability} Frequent retraining of the model comes at a price of its stability, i.e., giving the same prediction for the same input day in day out. To explain this phenomena we look at the changes in the rules that the model is learning in different daily runs. We generate a simplified approximation of our random forest model by producing a set of decision rules of a form $(origin, destination, airline)$, representing the cases when combination itineraries with a given $airline$ perform well on a given $(origin, destination)$ route. We analyse how many of the rules generated in day $T_{i-1}$ were dropped in the day $T_i$'s run of the model and how many new ones were added instead (Fig.~\ref{fig:temporal_stability}).

We see that around 88\% of rules remain relevant between the two consecutive days the remaining $\approx12\%$ are dropped and a similar number of new ones are added. Our qualitative investigation followed from this experiment suggested that dropping a large number of rules may end up in a negative user experience. Someone who saw a combination option on day $T_{i-1}$ might be frustrated from not seeing it on $T_i$ even if the price went up and it is no longer in the top ten of the search results. To account for this phenomenon we have introduced a simple heuristic in production which ensures that all of the rules which were generated on day $T_{i-1}$ will be included for another day $T_i$.

\subsection{Architecture of the pipeline}

\begin{figure}
\centering
\includegraphics[width = 0.8\columnwidth]{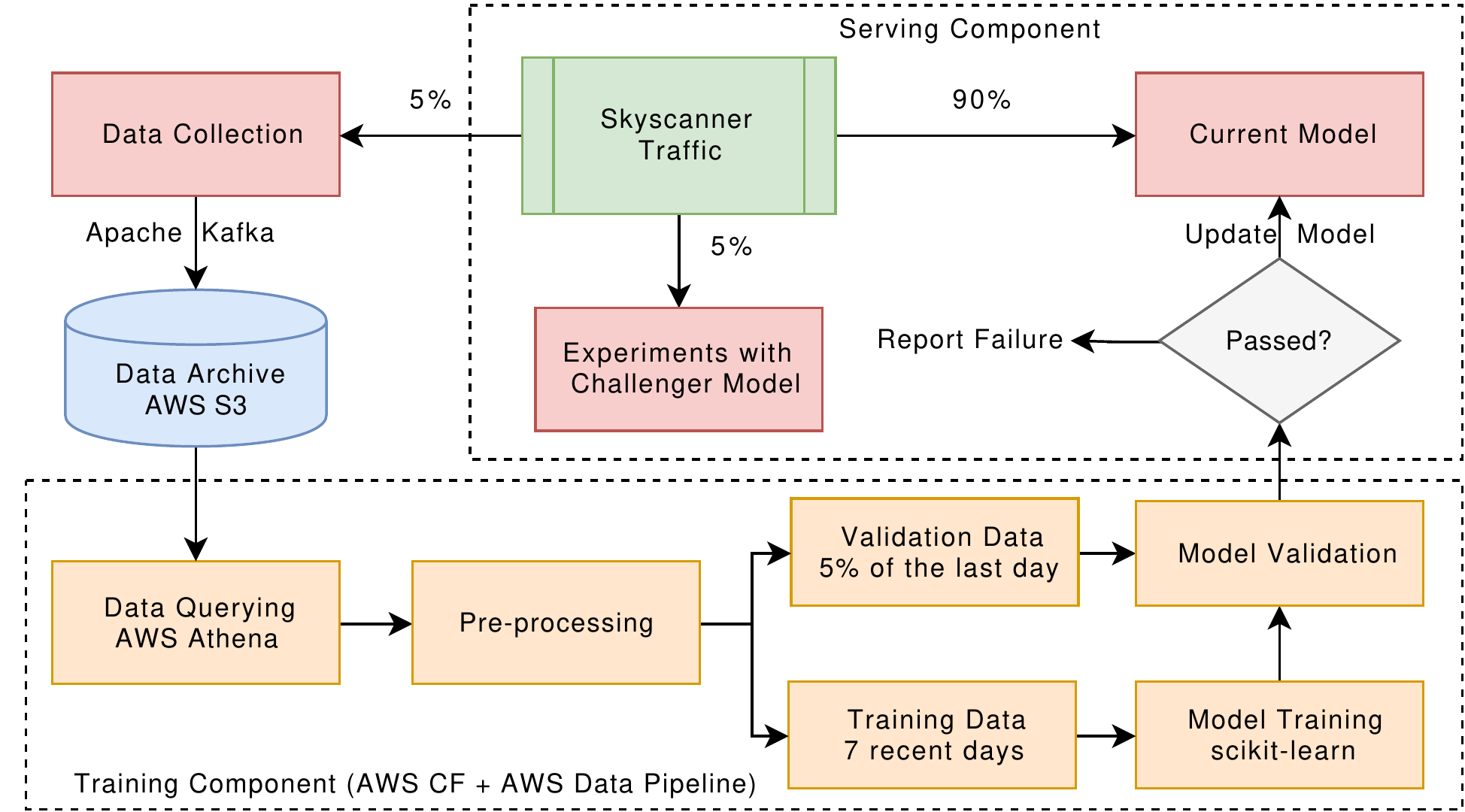}
\caption{The architecture of the machine learning pipeline.}
\label{fig:ml_pipeline}
\end{figure}

Equipped with the observations from the previous section we implement a machine learning pipeline summarised in Fig.~\ref{fig:ml_pipeline}. There are three main components in the design of the pipeline: the data collection process which samples the ground truth space to generate training data; the training component which runs daily to train and validate the model and the serving component which delivers predictions to the Skyscanner search engine.

\paragraph{Training infrastructure:} The training infrastructure is orchestrated by AWS Cloud Formation\footnote{https://aws.amazon.com/cloudformation/} and AWS Data Pipeline\footnote{https://aws.amazon.com/datapipeline/}. The data querying and preprocessing is implemented with Presto distributed computing framework\footnote{https://prestodb.io/} managed by AWS Athena\footnote{https://aws.amazon.com/athena/}. The model training is done with scikit-learn library on a high-capacity virtual machine. Our decision for opting towards a single large virtual machine vs. a multitude of small distributed ones has been dictated by the following considerations:\newline

\textbf{Data volume:} Once the heavy-lifting of data collection and preprocessing is done in Presto, the size of the resulting training data set becomes small enough to be processed on a single high capacity virtual machine.

\textbf{Performance:} By avoiding expensive IO operations characteristic of distributed frameworks, we decreased the duration of a model training cycle to less than 10 minutes.

\textbf{Technological risks:} The proposed production environment closely resembles our offline experimentation framework, considerably reducing the risk of a performance difference between the model developed during offline experimentation and the model run in production.

\paragraph{Traffic allocation} We use 5\% of Skyscanner search traffic to enable ground truth sampling and prepare the data set for training using Skyscanner's logging infrastructure\footnote{More details here https://www.youtube.com/watch?v=8z59a2KWRIQ} which is built on top of Apache Kafka\footnote{https://kafka.apache.org/}. We enable construction of all possible combination itineraries on this selected search traffic, collecting a representative sample of competitive and non-competitive cases to train the model. We use another ~5\% of the search traffic to run a challenger experiment when a potentially better performing candidate model is developed using offline analysis. The remaining 90\% of the search traffic are allocated to serve the currently best performing model.

\paragraph{Validation mechanism} We use the most recent seven days, $T_{i-7}..T_{i-1}$, of the ground truth data to train our model on day $T_i$ as explained in section \ref{sec:temporal}. We also conduct a set of validation tests on the newly trained model before releasing it to the serving infrastructure. We use a small share of the ground truth data (5\% out of 5\% of the sampled ground truth data) from the most recent day $T_{i-1}$ in the ground truth data set with the aim of having our validation data as close in time to when the model appears in production on day $T_i$. This sampled validation set is excluded from the training data.

\begin{figure}
\centering
\includegraphics[width = 0.65\columnwidth]{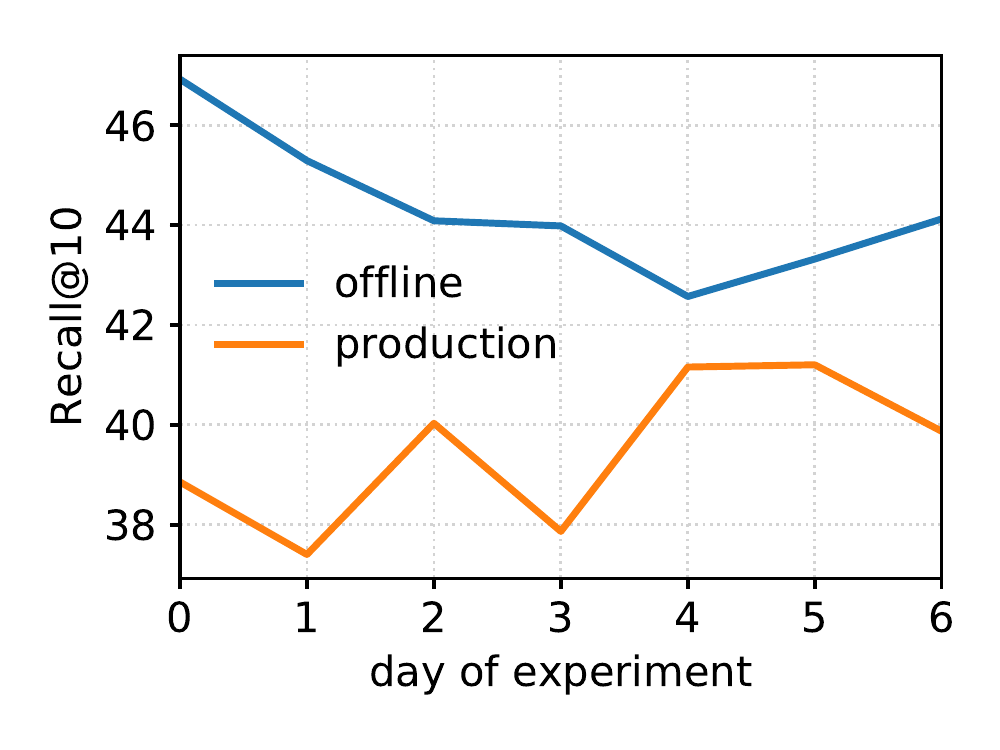}
\caption{Performance of the model in offline experiments vs. production expressed in terms of Recall@10 at 5\% of quote requests.}
\label{fig:recall_online}
\end{figure}

\subsection{Performance in production}

When serving the model in production we allow a budget of an additional 5\% of quote requests with which we expect to reconstruct 45\% of all competitive combination itineraries (recall Fig.~\ref{fig:performance_encodings}). From Fig.~\ref{fig:recall_online} we note that the recall measured in production deviates by $\approx 5\%$ from expectations in our offline experiments. We attribute this to model staleness incurred from 24 hour lag in the training data we use from the time when the model is pushed to serve users' searches.

Analysing the model's impact on Skyscanner users, we note that new cheap combination itineraries become available in 22\% of search results. We see evidence of users finding these additional itineraries useful with a 20\% relative increase in the booking transactions for combinations.

\section{Related work}
\label{sec:related_work}

\paragraph{Mining flights data}

The problem of airline fare prediction is discussed in detail in \cite{boyd2016future} and several data mining models were benchmarked in \cite{etzioni2003buy}. The authors of \cite{ayhan2016aircraft} modelled 3D trajectories of flights based on various weather and air traffic conditions. The problem of itinerary relevance ranking in one of the largest Global Distributed Systems was presented in \cite{mottini2017deep}. The systematic patterns of airline delays were analysed in \cite{fleurquin2013systemic}. And the impact of airport network structure on the spread of global pandemics was weighed up in \cite{colizza2006role}.

\paragraph{Location representation}

Traditional ways to model airline prices have been based on complex networks \cite{fleurquin2013systemic}\cite{colizza2006role} or various supervised machine learning models \cite{etzioni2003buy}\cite{mottini2017deep}. A more recent trend is around incorporating neural embeddings to model location data. Embeddings have seen great success in natural language processing~\cite{mikolov2013distributed}, modelling large graphs~\cite{perozzi2014deepwalk} and there has been a spike of enthusiasm around applying neural embedding to geographic location context with a variety of papers focusing on: a) mining embeddings from sequences of locations~\cite{pang2016deepcity}\cite{zhao2017geo}\cite{liu2016exploring}\cite{zhao2017geo}; b) modelling geographic context~\cite{yan2017itdl}\cite{feng2017poi2vec}\cite{kejriwal2017neural} and c) using alternative neural architectures where location representations are learned while optimising towards particular applications~\cite{yan2017itdl} and different approaches are mixed together in~\cite{kejriwal2017neural} and~\cite{feng2017poi2vec}. The practicalities of augmenting existing non-deep machine learning pipelines with neural embeddings are discussed in \cite{zhu2017deep} and in \cite{chamberlain2017customer}.

\paragraph{Productionising machine learning systems}

The research community has recently started recognising the importance of sharing experience and learning in the way machine learning and data mining systems are implemented in production systems. In \cite{sculley2015hidden} the authors stress the importance of investing considerable thinking and resources in building long-lasting technological infrastructures for machine learning systems. The authors of \cite{liu2017related} describe their experiences in building a recommendation engine, providing a great summary of business and technological constraints in which machine learning researchers and engineers operate when working on production systems. In \cite{tata2017quick} the developers of Google Drive share their experience on the importance of reconsidering UI metrics and launch strategies for online experimentation with new machine learning features. Alibaba research in \cite{liu2017cascade} emphasises the importance of considering performance constraints and user experience and feedback in addition to accuracy when deploying machine learning in production.

\section{Conclusions}

We have presented a system that learns to build cheap and novel round trip flight itineraries by combining legs from different airlines. We collected a sample of all such combinations and found that the majority of competitive combinations were concentrated around a minority of airlines but equally spread across routes of differing popularity. We also found that the performance of these combinations in search results increases as the time between search and departure date decreases.

We formulated the problem of predicting competitive itinerary combinations as a trade-off between the coverage in the search results and the cost associated with performing the requests to airlines for the quotes needed for their construction. We considered a variety of supervised learning approaches to model the proposed prediction problem and showed that richer representations of location data improved performance.

We put forward a number of practical considerations for putting the proposed model into production. We showed the importance of considering the trade-off between the model stability and staleness, balancing keeping the model performant whilst minimising the potential negative impact on the user experience that comes with changeable website behaviour.

We also identify various considerations we took to deliver proposed model to users including technological risks, computational complexity and costs. Finally, we provided an analysis of the model's performance in production and discuss its positive impact on Skyscanner's users.

\section*{Acknowledgement}
The authors would like to thank the rest of the Magpie team (Boris Mitrovic, Calum Leslie, James Eastwood, Linda Edstrand, Ronan Le Nagard, Steve Morley, Stewart McIntyre and Vitaly Khamidullin) for their help and support with this project and the following people for feedback on drafts of this paper: Bryan Dove, Craig McIntyre, Kieran McHugh, Lisa Imlach, Ruth Garcia, Sri Sri Perangur, Stuart Thomson and Tatia Engelmore.

\bibliographystyle{splncs04}

\begin{thebibliography}{10}
\providecommand{\url}[1]{\texttt{#1}}
\providecommand{\urlprefix}{URL }
\providecommand{\doi}[1]{https://doi.org/#1}

\bibitem{ayhan2016aircraft}
Ayhan, S., Samet, H.: Aircraft trajectory prediction made easy with predictive
  analytics. In: KDD. pp. 21--30 (2016)

\bibitem{boyd2016future}
Boyd, E.: The future of pricing: How airline ticket pricing has inspired a
  revolution. Springer (2016)

\bibitem{chamberlain2017customer}
Chamberlain, B.P., Cardoso, A., Liu, C.H., Pagliari, R., Deisenroth, M.P.:
  Customer life time value prediction using embeddings. In: Proceedings of the
  ninth ACM SIGKDD international conference on Knowledge discovery and data
  mining. ACM (2017)

\bibitem{colizza2006role}
Colizza, V., Barrat, A., Barth{\'e}lemy, M., Vespignani, A.: The role of the
  airline transportation network in the prediction and predictability of global
  epidemics. Proceedings of the National Academy of Sciences of the United
  States of America  \textbf{103}(7),  2015--2020 (2006)

\bibitem{etzioni2003buy}
Etzioni, O., Tuchinda, R., Knoblock, C.A., Yates, A.: To buy or not to buy:
  mining airfare data to minimize ticket purchase price. In: Proceedings of the
  ninth ACM SIGKDD international conference on Knowledge discovery and data
  mining. pp. 119--128. ACM (2003)

\bibitem{feng2017poi2vec}
Feng, S., Cong, G., An, B., Chee, Y.M.: Poi2vec: Geographical latent
  representation for predicting future visitors. In: AAAI. pp. 102--108 (2017)

\bibitem{fleurquin2013systemic}
Fleurquin, P., Ramasco, J.J., Eguiluz, V.M.: Systemic delay propagation in the
  us airport network. Scientific reports  \textbf{3}, ~1159 (2013)

\bibitem{guo2016entity}
Guo, C., Berkhahn, F.: Entity embeddings of categorical variables. arXiv
  preprint arXiv:1604.06737  (2016)

\bibitem{kejriwal2017neural}
Kejriwal, M., Szekely, P.: Neural embeddings for populated geonames locations.
  In: International Semantic Web Conference. pp. 139--146. Springer (2017)

\bibitem{liu2017related}
Liu, D.C., Rogers, S., Shiau, R., Kislyuk, D., Ma, K.C., Zhong, Z., Liu, J.,
  Jing, Y.: Related pins at pinterest: The evolution of a real-world
  recommender system. In: Proceedings of the 26th International Conference on
  World Wide Web Companion. pp. 583--592. International World Wide Web
  Conferences Steering Committee (2017)

\bibitem{liu2017cascade}
Liu, S., Xiao, F., Ou, W., Si, L.: Cascade ranking for operational e-commerce
  search. In: Proceedings of the 23rd ACM SIGKDD International Conference on
  Knowledge Discovery and Data Mining. pp. 1557--1565. ACM (2017)

\bibitem{liu2016exploring}
Liu, X., Liu, Y., Li, X.: Exploring the context of locations for personalized
  location recommendations. In: IJCAI. pp. 1188--1194 (2016)

\bibitem{mikolov2013distributed}
Mikolov, T., Sutskever, I., Chen, K., Corrado, G.S., Dean, J.: Distributed
  representations of words and phrases and their compositionality. In: Advances
  in neural information processing systems. pp. 3111--3119 (2013)

\bibitem{mottini2017deep}
Mottini, A., Acuna-Agost, R.: Deep choice model using pointer networks for
  airline itinerary prediction. In: Proceedings of the 23rd ACM SIGKDD
  International Conference on Knowledge Discovery and Data Mining. pp.
  1575--1583. ACM (2017)

\bibitem{pang2016deepcity}
Pang, J., Zhang, Y.: Deepcity: A feature learning framework for mining location
  check-ins. arXiv preprint arXiv:1610.03676  (2016)

\bibitem{perozzi2014deepwalk}
Perozzi, B., Al-Rfou, R., Skiena, S.: Deepwalk: Online learning of social
  representations. In: Proceedings of the 20th ACM SIGKDD international
  conference on Knowledge discovery and data mining. pp. 701--710. ACM (2014)

\bibitem{sculley2015hidden}
Sculley, D., Holt, G., Golovin, D., Davydov, E., Phillips, T., Ebner, D.,
  Chaudhary, V., Young, M., Crespo, J.F., Dennison, D.: Hidden technical debt
  in machine learning systems. In: Advances in Neural Information Processing
  Systems. pp. 2503--2511 (2015)

\bibitem{tata2017quick}
Tata, S., Popescul, A., Najork, M., Colagrosso, M., Gibbons, J., Green, A.,
  Mah, A., Smith, M., Garg, D., Meyer, C., et~al.: Quick access: Building a
  smart experience for google drive. In: Proceedings of the 23rd ACM SIGKDD
  International Conference on Knowledge Discovery and Data Mining. pp.
  1643--1651. ACM (2017)

\bibitem{yan2017itdl}
Yan, B., Janowicz, K., Mai, G., Gao, S.: From itdl to place2vec--reasoning
  about place type similarity and relatedness by learning embeddings from
  augmented spatial contexts. Proceedings of SIGSPATIAL  \textbf{17},  7--10
  (2017)

\bibitem{zhao2017geo}
Zhao, S., Zhao, T., King, I., Lyu, M.R.: Geo-teaser: Geo-temporal sequential
  embedding rank for point-of-interest recommendation. In: Proceedings of the
  26th International Conference on World Wide Web Companion. pp. 153--162.
  International World Wide Web Conferences Steering Committee (2017)

\bibitem{zhu2017deep}
Zhu, J., Shan, Y., Mao, J., Yu, D., Rahmanian, H., Zhang, Y.: Deep embedding
  forest: Forest-based serving with deep embedding features. In: Proceedings of
  the 23rd ACM SIGKDD International Conference on Knowledge Discovery and Data
  Mining. pp. 1703--1711. ACM (2017)

\end{thebibliography}

\end{document}